\pdfoutput=1
\documentclass[11pt]{article}

\usepackage{emnlp2021}

\usepackage{times}
\usepackage{latexsym}

\usepackage[T1]{fontenc}
\usepackage[utf8]{inputenc}

\usepackage{microtype}

\usepackage{CJKutf8}
\newcommand*{\ja}[1]{% 
    \begin{CJK}{UTF8}{ipxm}#1\end{CJK}}

\usepackage{graphicx}
\usepackage{tabularx}
\newcommand{\bhline}{\noalign{\hrule height 1.5pt}} % 太い横線の定義
\usepackage{amsmath}
\usepackage{multirow}
\usepackage{arydshln}
\usepackage{dashbox}
\title{Transformer-based Lexically Constrained Headline Generation}

\author{
  Kosuke Yamada$^{1}$\thanks{\ \ Work done during an internship at The Asahi Shimbun Company} \ \ \ \ \ \ \ 
  Yuta Hitomi$^{2}$\thanks{\ \ Work done at The Asahi Shimbun Company} \ \ \ \ \ \ \ 
  Hideaki Tamori$^{3}$ \ \ \ \ \ \ \ 
  Ryohei Sasano$^{1}$ \\
  {\bf Naoaki Okazaki$^{4}$ \ \ \ \ \ \ \ 
  Kentaro Inui$^{5,6}$ \ \ \ \ \ \ \ 
  Koichi Takeda$^{1}$} \\
  $^{1}$Nagoya University \ \ \ \ \ \ \ 
  $^{2}$Insight Edge, Inc. \ \ \ \ \ \ \ \ 
  $^{3}$The Asahi Shimbun Company \\
  $^{4}$Tokyo Institute of Technology \ \ \ \ \ \ \ \ 
  $^{5}$Tohoku University \ \ \ \ \ \ \ \ 
  $^{6}$RIKEN AIP \\
  {\tt yamada.kosuke@c.mbox.nagoya-u.ac.jp}, \\ 
  {\tt yuta.hitomi@insightedge.jp}, \
  {\tt tamori-h@asahi.com}, \\
  {\tt \{sasano,takedasu\}@i.nagoya-u.ac.jp}, \\
  {\tt okazaki@c.titech.ac.jp}, \
  {\tt inui@ecei.tohoku.ac.jp} \\
}

\begin{document}
\maketitle
\begin{abstract}
This paper explores a variant of automatic headline generation methods, where a generated headline is required to include a given phrase such as a company or a product name.
Previous methods using Transformer-based models generate a headline including a given phrase by providing the encoder with additional information corresponding to the given phrase. 
However, these methods cannot always include the phrase in the generated headline.
Inspired by previous RNN-based methods generating token sequences in backward and forward directions from the given phrase, we propose a simple Transformer-based method that guarantees to include the given phrase in the high-quality generated headline.
We also consider a new headline generation strategy that takes advantage of the controllable generation order of Transformer.
Our experiments with the Japanese News Corpus demonstrate that our methods, which are guaranteed to include the phrase in the generated headline, achieve ROUGE scores comparable to previous Transformer-based methods.
We also show that our generation strategy performs better than previous strategies.
\end{abstract}

\section{Introduction}
Following the initial work of \newcite{rush15}, abstractive headline generation using the encoder-decoder model has been studied extensively \cite{chopra2016,nallapati2016,paulus2018}.
In the automatic headline generation for advertising articles, there are requests to include a given phrase such as a company or product name in the headline.

Generating a headline that includes a given phrase has been considered one of the lexically constrained sentence generation tasks.
For these tasks, there are two major approaches.
One approach is to select a plausible sentence including the given phrase from several candidate sentences generated from left to right
\cite{hokamp2017,anderson2017,post2018}.
Although these methods can include multiple phrases in a generated sentence, they are computationally expensive due to the large search space of the decoding process. 
In addition, since they try to force given phrases into sentences at every step of the generation process, these methods may harm the quality of the generated sentence \cite{liu2019}.

Another approach proposed by \newcite{mou2015} is to generate token sequences in backward and forward directions from the given phrase.
\newcite{mou2016} proposed Sequence to Backward and Forward Sequences (Seq2BF), which applies the method of \newcite{mou2015} to the sequence-to-sequence (seq2seq) framework.
They use an RNN-based model and adopt the best strategies proposed by \newcite{mou2015}, generating the backward sequence from the phrase and then generating the remaining forward sequence.
\newcite{liu2019} introduced the Generative Adversarial Network (GAN) to the model of \newcite{mou2015} to resolve the exposure bias problem \cite{bengio2015} caused by generating sequences individually, and used the attention mechanism \cite{bahdanau2015} to improve the consistency between both sequences.
However, their model does not support the seq2seq framework.

Recently, \newcite{he2020} used a Transformer-based model \cite{vaswani17}, which is reported to achieve high performance, to generate a headline containing a given phrase.
They proposed providing an encoder with additional information related to the given phrase.
However, their method may not always include the given phrases in the generated headline.

In this study, we work on generating lexically constrained headlines using Transformer-based Seq2BF.
The RNN-based model used by \newcite{mou2016} executes a strategy of continuous generation in one direction, and thus cannot utilize the information of the forward sequence when generating the backward sequence.
However, Transformer can execute a variety of generative strategies by devising attention masks, so it can solve the problem of the RNN-based model.
We propose a new strategy that generates each token from a given phrase alternately in the backward and forward directions, in addition to adapting and extending the strategies of \newcite{mou2016} to the Transformer architecture.

Our experiments with a Japanese summarization corpus show that our proposed method always includes the given phrase in the generated headline and achieves performance comparable to previous Transformer-based methods.
We also show that our proposed generating strategy performs better than the extended strategy of the previous methods.

\section{Proposed Method}
\begin{figure}[!t]
\centering
\includegraphics[width=0.85\linewidth]{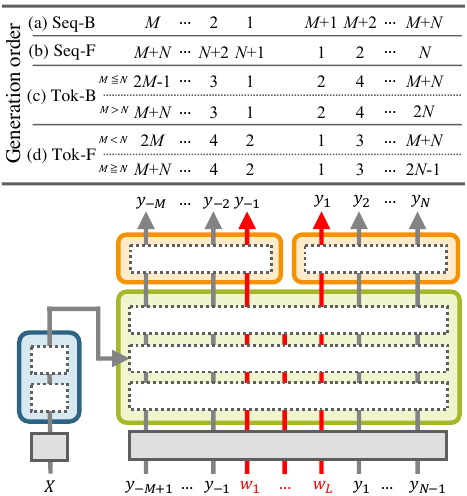}
\caption{Overview of Transformer-based Seq2BF. 
Blue, green, and orange boxes indicate Transformer encoder, decoder, and generation components, respectively.
The arrow from the encoder to the decoder represents that the decoder's attention mechanism refers to the output from the encoder.}
\label{fig:method}
\end{figure}

We propose a Transformer-based Seq2BF model that applies Seq2BF proposed by \newcite{mou2016} to the Transformer model to generate headlines including a given phrase.
The Seq2BF takes $W (=w_1,...,w_L;w_{1:L})$ as the given phrase consisting of $L$ tokens and generates the headline $y_{-M:-1}$ of $M$ tokens backward from $W$, and the headline $y_{1:N}$ of $N$ tokens forward from $W$.
The Transformer-based Seq2BF is the Transformer model with two generation components, consisting of a linear and a softmax layer (see Figure \ref{fig:method}).

In Transformer-based Seq2BF unlike Transformer generating tokens from left to right, the token position changes relatively depending on already generated tokens.
We determine the token position, inputting to the positional encoding layer of the decoder, $\lfloor\frac{L+1}{2}\rfloor$ in $W$ to be 0, and the position in the backward direction to be negative, and the position in the forward direction to be positive.

We consider the following four generation strategies.
In addition to two strategies (a) and (b), which extend those proposed by \newcite{mou2016}, we proposfe new strategies (c) and (d) as step-wise alternating generation to keep better contextual consistency in both backward and forward directions.
\begin{description}
\setlength{\itemsep}{-0.5ex}
\item[\textmd{(a)}] Generating a \textbf{sequence} \textbf{backward} and then a sequence forward. (\textbf{Seq-B})
\item[\textmd{(b)}] Generating a \textbf{sequence} \textbf{forward} and then a sequence backward. (\textbf{Seq-F})
\item[\textmd{(c)}] Generating each \textbf{token} \textbf{backward} and then forward alternately. (\textbf{Tok-B})
\item[\textmd{(d)}] Generating each \textbf{token} \textbf{forward} and then backward alternately. (\textbf{Tok-F})
\end{description}

\noindent Transformer-based Seq2BF is formulated as
\begin{equation}
P(Y|X, W) = \displaystyle{\prod_{j \in POS_j}}  \ P(y_j| Y_{obs}, X), \\
\label{eq:dog}
\end{equation}

\noindent where $X$ denotes tokens of the article, $W$ denotes tokens of the given phrase, $Y (=y_{-M:-1},w_{1:L},y_{1:N})$ denotes tokens of the final generated headline, and $Y_{obs}$ denotes the already-generated partial headline including $W$. 
Also, $POS_j$ denotes a list of token positions representing the order of tokens to be generated corresponding to each generation strategy (see Figure 1), for example $[-1,-2,...,-M,1,2,...,N]$ in Seq-B.
In Tok-B/F which $M$ and $N$ are different, once the generation in one direction is completed, the generation will be continued only in the remaining directions until $M+N$ steps.
For example in the case of $M>N$ in Tok-B, our method completes generating tokens in the forward direction first, so it generates them in both directions until the $2N$ step, and then generates them only in the backward direction from the $2N+1$ step to the $M+N$ step.

\begin{figure*}[!t]
\centering
\includegraphics[width=0.8\linewidth]{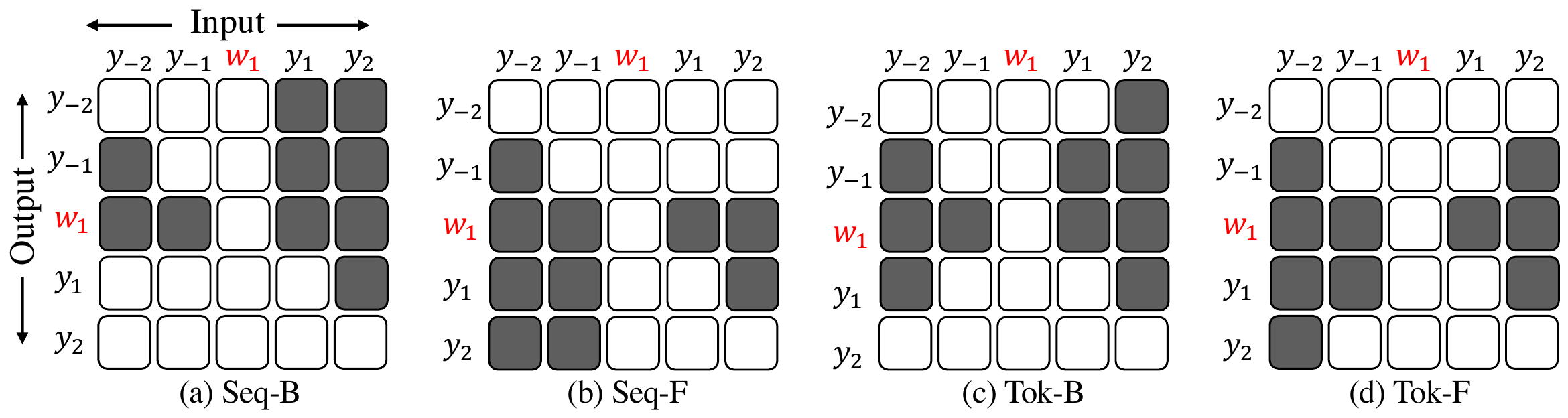}
\caption{Attention mask patterns on the decoder side during training, used for each generation strategy in Transformer-based Seq2BF. 
The dark cells indicate the masked areas.
These are examples of the headline with the length of 5, where $w_1$ is specified as the phrase.}
\label{fig:mask}
\end{figure*}

To train the model on these generative strategies, we have prepared an attention mask for the decoder.
Transformer can control the generation order of tokens by devising the attention mask used in the decoder's self-attention mechanism.
Transformer generates tokens from left to right, so it is sufficient to disable the attention to tokens forward from the input tokens.
However, the Transformer-based Seq2BF needs to specify the areas where input tokens disallow the attention in the backward and forward directions, depending on each generation strategy (see Figure \ref{fig:mask}).

\section{Experiment}
We conducted the experiment to verify the performance of our methods in the headline generation task.
The objective of our experiment is to compare our method with previous Transformer-based methods that generate tokens from left to right.
We also compare Seq-B/F, the generation orders proposed by \newcite{mou2016}, with Tok-B/F, our new generation orders.

\subsection{Setting}
We used the 2019 version of the Japanese News Corpus (JNC)\footnote{\url{https://cl.asahi.com/api_data/jnc-jamul-en.html}} \cite{hitomi19} as the dataset. 
The JNC contains 1,932,399 article-headline pairs, and we split them randomly at a ratio of 98:1:1 for use as training, validation, and test sets, respectively.\footnote{We applied the preprocessing script at \url{{https://github.com/asahi-research/script-for-transformer-based-seq2bf}} to the original JNC to obtain the split dataset.}
We utilized MeCab \cite{kudo2004} with the IPAdic\footnote{\url{https://taku910.github.io/mecab/}} and then applied the Byte Pair Encoding (BPE) algorithm\footnote{\url{https://github.com/rsennrich/subword-nmt}} \cite{gage1994} for tokenization.
We trained BPE with 10,000 merge operations and obtained the most frequent 32,000 tokens from the articles and the headlines, respectively.

We used context word sequences extracted from the reference headlines by GiNZA\footnote{\url{https://github.com/megagonlabs/ginza}} as the `given' phrase.\footnote{We used \textsf{ ``ginza.bunsetu\_phrase\_spans''} API.}
An average of 4.99 phrases was extracted from the reference headlines, and the `given' phrases consisted of an average of 2.32 tokens.
We evaluated our methods using precision, recall, and F-score of ROUGE-1/2/L \cite{lin04} and success rate (SR), which is the percentage of the headline that includes the given phrase.
We also calculated the Average Length Difference (ALD) to analyze the length of the generated headlines, as
\begin{equation}
\mbox{ALD} = \frac{1}{n}\sum_{i=1}^n l_i - \mbox{len}_i,
\label{eq:ale}
\end{equation}
where $n$, $l_i$, and $\mbox{len}_i$ are the number of samples, the length of the generated headline, and the length of the reference headline, respectively. 

\begin{table*}[t!]
\centering
\small
\begin{tabular}{l@{\ \ }lrccccc}
\bhline
&&  \multicolumn{1}{c}{\multirow{2}{*}{SR}} & ROUGE-1 & ROUGE-2 & ROUGE-L & \multicolumn{1}{c}{\multirow{2}{*}{ALD}} & \multicolumn{1}{c}{params} \\
&&     & P/R/F & P/R/F & P/R/F & {} & \multicolumn{1}{c}{$\times 10^6$} \\
\hline
\multicolumn{2}{l}{Transformer \cite{vaswani17}}                            & 36.3 & 57.1/48.9/51.4 & 29.8/25.2/26.5 & 47.1/40.9/42.8 & --3.62 & 72 \\ \hline
\multicolumn{2}{l}{Transformer version of \newcite{he2020}}       & 90.2 & 63.1/\textbf{54.8}/57.2 & 36.0/30.7/32.2 & 51.9/\textbf{45.4}/47.4 & --3.02 & 72 \\ \hline
&(Seq-B) & \textbf{100.0} & 63.6/52.4/55.8 &   37.4/30.2/32.3 &   54.3/44.2/47.5 & --4.19 & 80 \\
\multirow{2}{*}{Transformer-based Seq2BF}&(Seq-F) & \textbf{100.0} & 64.6/53.2/56.7 &   38.1/\textbf{30.8}/32.9 &   54.8/44.9/48.1 & --4.30 & 80 \\
&(Tok-B) & \textbf{100.0} & 66.6/52.9/\textbf{57.6} &   39.3/\textbf{30.8}/\textbf{33.6} &   55.6/45.0/\textbf{48.6} & --5.29 & 80 \\
&(Tok-F) & \textbf{100.0} & \textbf{67.6}/51.6/57.1 &   \textbf{40.1}/30.2/33.5 &   \textbf{56.7}/44.2/\textbf{48.6} & --6.05 & 80 \\
\bhline
\end{tabular}
\caption{Experimental results. SR means success rate, and P/R/F means Precision/Recall/F-score.}
\label{tab:results}
\end{table*}

\begin{figure*}[!t]
\centering
\includegraphics[width=0.95\linewidth]{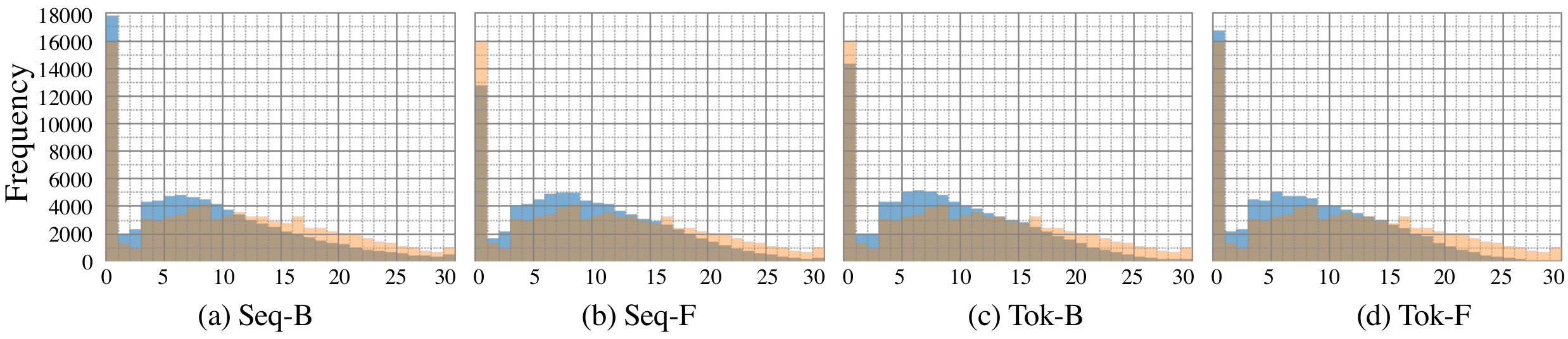}
\caption{Histogram of the character-level position of the given phrase in the headlines generated by Transformer-based Seq2BF.
Blue and orange bars indicate the generated and reference headlines, respectively.}
\label{fig:query_pos}
\end{figure*}

As a comparison method, we adopted the method proposed by \newcite{he2020} with vanilla Transformer instead of BART \cite{lewis2020}. 
This method controls the output by inserting the given phrase and the special token `|' in front of the input articles and randomly drops the given phrase from the input articles during training to improve the performance. 
The hyperparameters of both the comparison and our models are determined as described in \newcite{vaswani17}. 
The training was terminated when the perplexity computed on the validation set did not update three times in a row, and we used the model with the minimum perplexity on the validation set. 
The beam size during the inference was set to three.

\subsection{Results}
Table \ref{tab:results} shows the experimental results. 
Note that the proposed and compared methods achieved higher ROUGE scores than Transformer because we computed ROUGE scores between the reference and the system-generated headlines, including the phrase extracted from the reference headlines.

\begin{table*}[t!]
\small
\centering
\begin{tabularx}{\linewidth}{@{\ }c@{\ \ }|l@{\ }|c@{\ \ }|l@{\ }}
\bhline
\multicolumn{4}{X}{\textbf{Article:}
\ja{約1万匹のイワシが群れで泳ぐ様子を見られる京都水族館。この展示に合わせ、ちょっぴり変わった特別スイーツが6月末まで販売される。名前は「桜といわしのパフェ」で、...}}\\ 
\multicolumn{4}{X}{At the Kyoto Aquarium, you can see about 10,000 sardines swimming in schools.
To coincide with this exhibition, a special sweet that is slightly unique will be on sale until the end of June.
It is called ``Cherry Blossom and Sardine Parfait,'' and ...} \\ \hline
\multicolumn{4}{X}{\textbf{Reference Headline:} \ja{「目からウロコのおいしさ」\ \ \ 京都水族館にイワシパフェ}} \\ 
\multicolumn{4}{X}{``Scales Falling from Your Eyes'' -- Kyoto Aquarium Serves Sardine Parfait} \\ \hline
& \ja{\textcolor{red}{桜といわしのパフェ}} && \ja{「\textcolor{red}{桜といわしのパフェ}」特別スイーツ\ \ \ 京都水族館} \\ 
& \textcolor{red}{Cherry Blossom and} && Special Sweets ``\textcolor{red}{Cherry Blossom and Sardine Parfait}''\\ 
& \textcolor{red}{Sardine Parfait} && at Kyoto Aquarium \\ \cdashline{2-2}[5pt/5pt] \cdashline{4-4}[5pt/5pt]
\multirow{2}{*}{\textbf{Given}} & \ja{\textcolor{red}{6月末}} & \multirow{2}{*}{\textbf{Generated}} &
\ja{イワシの特別スイーツ、\textcolor{red}{6月末}まで販売 京都水族館} \\
\multirow{2}{*}{\textbf{Phrases}} & \textcolor{red}{the End of June} & \multirow{2}{*}{\textbf{Headlines}} & Special Sardine Sweets on Sale at Kyoto Aquarium until \textcolor{red}{the End of June} \\ \cdashline{2-2}[5pt/5pt] \cdashline{4-4}[5pt/5pt]
& \ja{\textcolor{red}{群れ}} &&
\ja{イワシの\textcolor{red}{群れ}のイワシ、特別スイーツに 京都水族館} \\
& \textcolor{red}{Schools} && Sardines of \textcolor{red}{Schools} of Sardines, to Special Sweets at Kyoto Aquarium \\ \cdashline{2-2}[5pt/5pt] \cdashline{4-4}[5pt/5pt]
& \ja{\textcolor{red}{約1万匹}} &&
\ja{イワシ\textcolor{red}{約1万匹}の特別スイーツ 京都水族館} \\ 
& \textcolor{red}{About 10,000} && Special Sweets of \textcolor{red}{About 10,000} Sardines at Kyoto Aquarium \\
\bhline
\end{tabularx}
\caption{Examples of headlines generated by Transformer-based Seq2BF (Tok-B).}
\label{tab:generation}
\end{table*}

Our methods always include the given phrase in the generated headlines, whereas the comparison method had a success rate of around 90\%. 
Although the recall of ROUGE scores tended to be higher in the comparison method than in the proposed method, the precision and F-scores of ROUGE scores in the proposed method were comparable or higher than in the comparison method.
As we notice from ALD, we found that Transformer-based Seq2BF generated shorter headlines than the Transformer models. 
It has been confirmed that the Transformer models with a single output direction tend to generate shorter headlines than the reference.
Because Transformer-based Seq2BF has two output directions, the generated headlines were considered to be even shorter.
This is the reason why our methods had lower recall scores than the comparison methods. 
Comparing the generation strategies of Transformer-based Seq2BF, we can see that Tok-B/F had a higher score than Seq-B/F.

To analyze how the four generation strategies of Transformer-based Seq2BF affected the system-generated headlines, we showed the character-level position of the given phrase in the headline using histograms in Figure \ref{fig:query_pos}.
As we can see, all generation strategies had similar distributions in the reference and system-generated headlines, and hence Transformer-based Seq2BF has also been presumed to learn the position of a given phrase in the headline.
Focusing on the headlines that include the given phrase in the head, the difference between the reference and the headline generated by Tok-B/F is smaller than that of the headline generated by Seq-B/F.
Also, the headlines generated by Seq-B tend to place the given phrase in the beginning, while this tendency is opposite for the headlines generated by Seq-F.

Table \ref{tab:generation} shows examples of the headlines generated by the Transformer-based Seq2BF (Tok-B).
When a product name such as \ja{``桜とイワシのパフェ''} (``Cherry Blossom and Sardine Parfait'') was given, our methods could generate a natural headline that includes the given phrase. 
Also, given the phrase \ja{``6月末''} (``the End of June''), our methods generated a headline with the addition of \ja{``販売''} (``on Sale'') that matched the given phrase.
On the other hand, we found the problem of generating the same words related to the given phrase in the backward and forward directions, such as the headline generated given \ja{``群れ''} (``Schools'').
In addition, given the phrase \ja{``約1万匹''} (``About 10,000''), our methods generated the headline meaning that special sweets contain about 10,000 sardines.
In this way, examples that were not faithful to the article were confirmed.

As can be seen from Table \ref{tab:generation}, various headlines are generated according to given phrases.
In general, it is difficult to control the diversity in headline generation, but our methods can generate diverse headlines by giving a variety of phrases.
However, it may be necessary to discuss whether our methods could generate diverse headlines.
The reason is that all examples are only partially diverse.
Specifically, they always include \ja{``特別スイーツ''} (``Special Sweets'') and \ja{``京都水族館''} (``Kyoto Aquarium'') as important contents in the headline.

\section{Conclusion}
We proposed Transformer-based Seq2BF that generates the lexically constrained headline by devising the attention mask for the decoder and generating backward and forward sentences from the phrase.
Our experiments using the JNC demonstrated that Transformer-based Seq2BF always includes the given phrase in the generated headline and obtains comparable performance compared to previous Transformer-based methods.
We also showed that strategies of generating each token alternately between backward and forward directions are more effective than that of generating a sequence in one direction and then a sequence in another direction.

In future work, we will investigate whether Transformer-based Seq2BF can generate natural headlines even when given a variety of phrases, such as phrases not in the reference or the article, and examine if our methods can creatively generate diverse headlines by giving a variety of phrases quantitatively.
Also, we will explore methods for generating headlines that include multiple phrases.

\bibliography{emnlp2021}
\bibliographystyle{acl_natbib}

% \appendix

% \section{Example Appendix}
% \label{sec:appendix}

% This is an appendix.

\end{document}